\documentclass[letterpaper, 10 pt, conference]{ieeeconf}  

\IEEEoverridecommandlockouts                              

\usepackage{graphics} 
\usepackage{epsfig} 
\usepackage{amsmath} 
\usepackage{amssymb}  
\usepackage{mathtools}

\newcommand{\methodname}{SAMBAR}

\let\labelindent\relax 
\usepackage{enumitem}
\usepackage{booktabs}
\usepackage{makecell}
 \usepackage{algorithm}
 \usepackage{algpseudocode}
 \usepackage{etoolbox}
\makeatletter
\apptocmd{\fs@ruled}{%
  \preto{\@fs@pre}{\vskip 4pt}%
}{}{}
\makeatother

\usepackage[dvipsnames]{xcolor}
\usepackage[font=footnot
esize]{caption}
\definecolor{taskorange}{HTML}{E46C0B}
\definecolor{taskblue}{HTML}{548ED5}
\definecolor{taskgreen}{HTML}{78933C}
\definecolor{mygray}{HTML}{7F7F7F}

\makeatletter
\long\def\@makecaption#1#2{%
\ifx\@captype\@IEEEtablestring%
\vskip\abovecaptionskip\relax%
\setbox\@tempboxa\hbox{\footnotesize #1: #2}%
\ifdim \wd\@tempboxa >\hsize%
\setbox\@tempboxa\hbox{\footnotesize #1: }%
\parbox[t]{\hsize}{\footnotesize \noindent\unhbox\@tempboxa#2}%
\else%
\hbox to\hsize{\footnotesize\hfil\box\@tempboxa\hfil}%
\fi%
\else
\@IEEEfigurecaptionsepspace%
\setbox\@tempboxa\hbox{\footnotesize #1.~~ #2}%
\ifdim \wd\@tempboxa >\hsize%
\setbox\@tempboxa\hbox{\footnotesize #1.~~ }%
\parbox[t]{\hsize}{\footnotesize \noindent\unhbox\@tempboxa#2}%
\else%
\ifcenterfigcaptions \hbox to\hsize{\footnotesize\hfil\box\@tempboxa\hfil}%
\else \hbox to\hsize{\footnotesize\box\@tempboxa\hfil}%
\fi\fi\fi}
\makeatother
\definecolor{mycitecolor}{RGB}{71, 191, 38}
\definecolor{mylinkcolor}{RGB}{40, 115, 201}
\makeatletter
\let\NAT@parse\undefined
\makeatother
\usepackage[bookmarks=true, colorlinks, citecolor=mycitecolor,linkcolor=mylinkcolor,urlcolor=mycitecolor]{hyperref}

\title{\LARGE \bf
\methodname: Selective Anchoring via Method of Multipliers \\ for Balanced Knowledge Acquisition and Retention \\ in Vision-Language-Action Models}

\author{Aayushi Shrivastava, Xunlan Zhou, Hongrui Zhao, Ziyu Chen, and Negar Mehr
\thanks{*This work is supported by the National Science Foundation, under grants ECCS-2438314 CAREER Award, CNS-2529645. This work was also supported in part by an Office of Naval Research (ONR) Young Investigator Program (YIP) award. This research was made possible by GPU resources provided via the NVIDIA Academic Grant.}
\thanks{ All authors are with the Department of Mechanical Engineering, University of California Berkeley, Berkeley, CA 94709, USA
        {\tt\small aayushis, wyattzhouxl, hongrui, ziyu, negar@berkeley.edu}}%
}

\begin{document}
\bstctlcite{IEEEexample:BSTcontrol}

\maketitle
\thispagestyle{empty}
\pagestyle{empty}

\begin{abstract}
Vision-Language-Action (VLA) models leverage large-scale pretraining to ultimately achieve generalist manipulation. 
Deployed VLA policies must support continual learning to acquire new tasks over time. Teaching a VLA a new task generally requires finetuning it on demonstrations of that task.
However, naively finetuning on downstream tasks causes the policy to forget earlier tasks and degrades generalist capabilities. This failure is known as catastrophic forgetting. Most continual learning methods counter it by replaying data from earlier tasks. However, the old task demonstrations are not always readily available.
In this paper, we introduce \methodname, a continual learning algorithm that prevents catastrophic forgetting during VLA finetuning without requiring access to the demonstrations of any previously learned task. We propose to cast continual learning as a constrained optimization problem and solve it with the method of multipliers. In our approach, the method of multipliers drives the policy to learn the new task without the model parameters drifting far away from their previous values. 
In contrast to a standard regularization penalty,  the method of multipliers raises the penalty as the constraint violation accumulates by using a dual variable. 
We also selectively anchor\footnote{Refers to keep model parameters near pretrained checkpoints, not LLM prompt/context anchoring} the parameters critical to previous tasks to preserve past knowledge, leaving other parameters free for new task acquisition.
The combination of dual variable and selective anchoring, therefore, balances knowledge acquisition with knowledge retention. 
We evaluate our method, \methodname, on the LIBERO simulation benchmark and on hardware. When sequentially finetuning on a VLA, every replay-free baseline we compare against completely forgets the first task it learned, whereas \methodname\ retains every task it has learned. Project website is \url{https://iconlab.negarmehr.com/SAMBAR/}.
\end{abstract}

\section{INTRODUCTION}\label{sec:introduction}
\begin{figure}[t]
    \centering
    \includegraphics[width=0.5\textwidth]{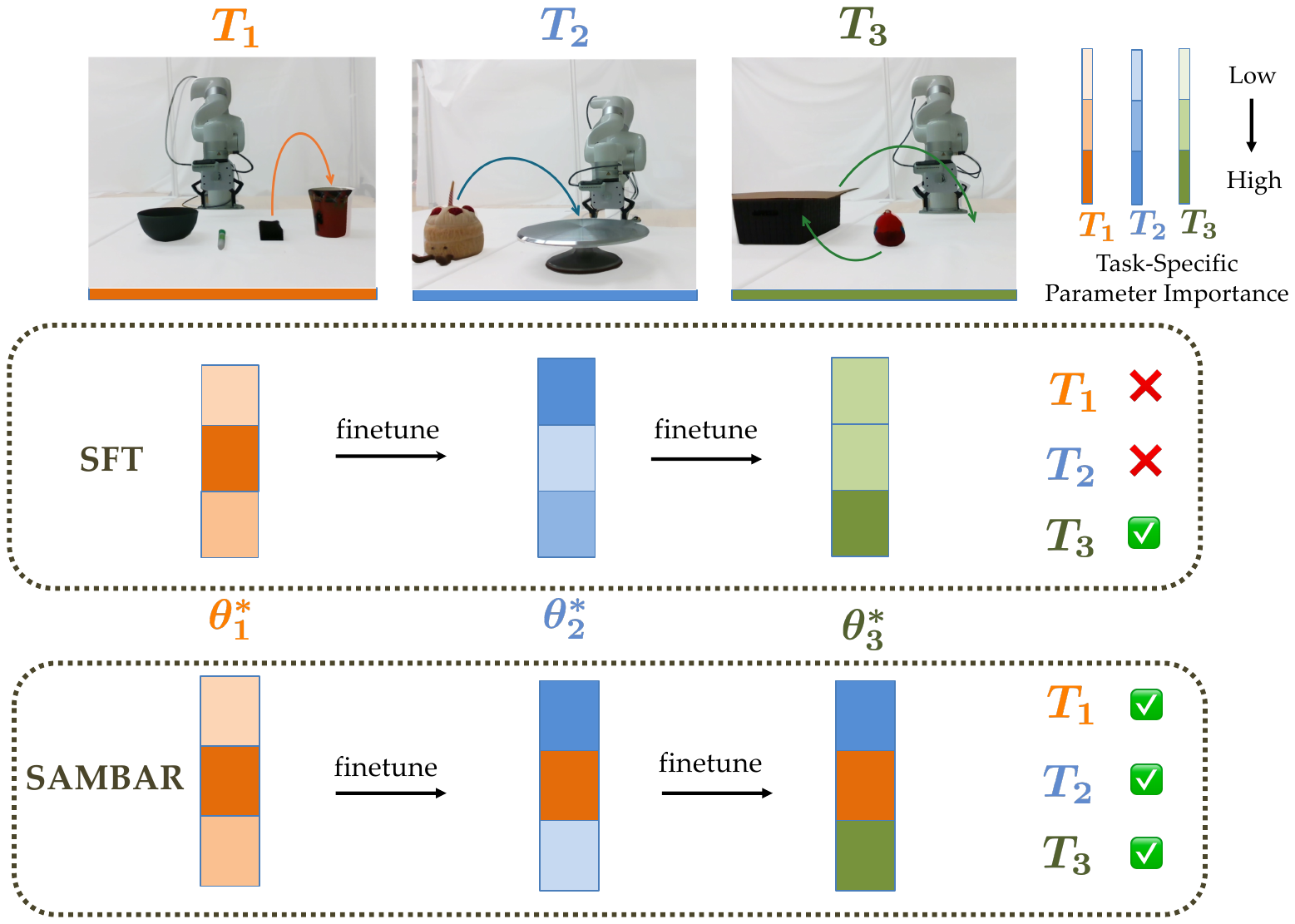}
    \vspace{-20pt}
    \caption{
\textbf{Overview of \methodname\ for replay-free continual VLA learning.}
A policy learns tasks \(T_1,T_2,T_3\) sequentially using only the current task's demonstrations.
\textcolor{taskorange}{Orange},
\textcolor{taskblue}{blue}, and
\textcolor{taskgreen}{green}
indicate parameter importance for
\(\textcolor{taskorange}{T_1}\),
\(\textcolor{taskblue}{T_2}\), and
\(\textcolor{taskgreen}{T_3}\), respectively,
with darker shades indicating greater importance.
Standard supervised finetuning (SFT) can overwrite parameters important to earlier tasks, causing forgetting.
\methodname\ combines an importance-weighted parameter anchor with the method of multipliers to limit changes to parameters important to prior tasks while allowing new-task adaptation.
}
    \label{fig:overview}
    \vspace{-20pt}
\end{figure}

Picture this: you bring home a brand-new assistant robot running a Vision-Language-Action (VLA) policy ~\cite{rt2_2023, openvla_2024, pi05_2025, molmoact_2025} that effortlessly does laundry and washes dishes. 
But then you decide you want it to clean your aquarium too.
Lacking the massive data and compute from its original pretraining, you may simply decide to finetune the VLA on a few demonstrations of aquarium tank cleaning. As a result, your robot ends up mastering the aquarium cleaning, only to completely forget how to wash dishes. 
This is a clear case of \emph{continual learning} running into \emph{catastrophic forgetting}. What we want instead is a policy that can take on a new task without losing the ones it already has.

Catastrophic forgetting is very common while training a neural network on tasks in sequence~\cite{kirkpatrick_overcoming_2017, langeContinualLearningSurvey2021}. Forgetting arises because under Supervised Finetuning (SFT)~\cite{bain_framework_1995}, the training objective that teaches the robot to clean the aquarium carries no term to protect the knowledge of washing dishes. 
One common solution to address this is to use a replay which combines demonstrations of the earlier tasks with the new ones during finetuning~\cite{lopezpaz_gradient_2017, chaudhry_tiny_2019, wan_lotus_2024}. 
However, we do not always have access to the demonstration data of all the prior tasks the VLA knows.
The company never shipped the demonstrations for folding laundry and washing dishes with the robot assistant. 

Another common category of work to avoid catastrophic forgetting is regularization-based methods. 
These methods often add a penalty term to the training objective to penalize the policy parameters from getting too far from what their prior values. 
However, tuning this penalty to balance learning new tasks while preventing catastrophic forgetting proves challenging in practice. 
Set it high and the robot never learns to clean the aquarium; set it low and it forgets how to wash dishes.
We thus ask: \emph{Can we develop a continual learning method that eliminates the need for previous training data while balancing knowledge acquisition and retention?}

In this paper, we introduce \textbf{\methodname} (\textbf{S}elective \textbf{A}nchoring via \textbf{M}ethod of Multipliers for \textbf{B}alanced Knowledge \textbf{A}cquisition and \textbf{R}etention in Vision-Language-Action Models), a replay-free continual learning algorithm that lets a VLA policy learn new tasks sequentially without forgetting the ones it already knows how to perform. Inspired by recent work~\cite{zhouTACOTemporalConsensus2026}, we propose to cast the continual learning problem as a constrained optimization problem where the objective is to learn a new task subject to the constraint that policy parameters do not drift far from their previous values, as shown in Figure~\ref{fig:overview}. We solve this constrained optimization problem with the method of multipliers~\cite{boyd_distributed_2011}. The method of multipliers accumulates the constraint violation in a dual variable and raises the price of violating the constraint until the policy satisfies it. 

We define the constraint as a weighted one: it regulates how far each parameter may drift from its previous value, weighted by an importance estimate of how much that parameter matters to the tasks the policy has already learned and to the new one it must learn. Parameters the earlier tasks rely on stay close to their previous values, while the rest are free to move and learn the new task as shown in Figure~\ref{fig:overview}. We evaluate \methodname\ on the LIBERO benchmark~\cite{libero_2023} against five replay-free baselines and demonstrate its effectiveness at learning new task sequentially on an xArm manipulator.

In summary, our contributions in this work are as follows:
\begin{enumerate}[itemsep=0pt, topsep=2pt, parsep=0pt, partopsep=0pt]
    \item We propose SAMBAR, a replay-free continual learning algorithm which finetunes VLAs using the method of multipliers with an importance weighting algorithm that prevents catastrophic forgetting even in a sequential finetuning setting. 
    \item We compare \methodname\ against five replay-free baselines on the LIBERO benchmark in a variety of settings. After finetuning on three tasks in sequence, every baseline has fallen to $0.0\%$ on the first task it was taught, while \methodname\ still performs it $84.0\%$ of the time and keeps $88.2\%$ success rate on the pretraining tasks.
    \item We demonstrate our method's effectiveness for learning new tasks in a sequential fashion on a robot arm.
\end{enumerate}

The remainder of this paper is organized as follows: In Section~\ref{sec:related_works}, we review related works. In Section~\ref{sec:problem}, we provide the problem formulation. In Section~\ref{sec:prelims}, we introduce our preliminaries. In Section~\ref{sec:method}, we explain the details of our proposed algorithm. In Section~\ref{sec:sim_results}, we report our results on the LIBERO benchmark and compare our method with relevant baselines. In Section~\ref{sec:hardware_results}, we present our results in hardware experiments. Finally, we conclude the paper and discuss future directions in Section~\ref{sec:conclusion}.

\section{RELATED WORKS}\label{sec:related_works}
Vision-language-action (VLA) models map visual observations and language instructions to actions, typically through large-scale imitation learning. RT-2~\cite{rt2_2023} and OpenVLA~\cite{openvla_2024} generate discrete action tokens autoregressively, while pi0.5~\cite{pi05_2025} uses a continuous action expert. Other models incorporate auxiliary supervision, such as depth-aware perception tokens~\cite{molmoact_2025}. After deployment, VLA policies may encounter new tasks beyond the task they already know, and must learn them without losing previously acquired skills. This motivates continual learning, which studies sequential task acquisition while retaining earlier knowledge~\cite{kirkpatrick_overcoming_2017, langeContinualLearningSurvey2021}. We review four approaches: replay, parameter isolation, model merging, and regularization.

The most common approach is replay: these methods keep a buffer of earlier data and mix it into the new task's training, either by finetuning on the two together~\cite{chaudhry_tiny_2019} or by rejecting updates that would raise the loss on the buffer~\cite{lopezpaz_gradient_2017}. Robot learning does the same through co-training~\cite{bethune_scaling_2025} and skill libraries~\cite{wan_lotus_2024}. Replay transfers to VLAs well: a small buffer drives a pretrained VLA's forgetting nearly to zero~\cite{liuPretrainedVisionLanguageActionModels2026}, and removing it leads to severe forgetting~\cite{zhu_can_2026}. But a deployed policy rarely has those demonstrations. We therefore work in the replay-free setting.

Among replay-free methods,  parameter isolation methods avoids interference by giving each task its own parameters. These methods freeze a pruned subnetwork~\cite{mallya_packnet_2018}, restrict updates to a subspace orthogonal to earlier gradients~\cite{farajtabar_orthogonal_2020}, or learn one adapter per skill~\cite{lora_2022, liu_skill_2025}. 
When applied to VLA models,  parameter isolation keeps the pretrained model frozen beside a small adaptation module~\cite{guo_priorvla_2026, huang_breaking_2026}. 
A related line of works uses reinforcement rather than supervised finetuning to limit drift~\cite{liu_lifelongrft_2026, zeng_crlvla_2026, hu_simple_2026}. 
These methods grow with every new skill, or they need an environment and a reward. 
In comparison, \methodname\ learns from demonstrations alone and returns a single policy of fixed size.

Model merging combines separately finetuned models by interpolating their weights, which works because models finetuned from the same initialization stay in one linearly connected basin~\cite{frankle_linear_2020}. Interpolating a finetuned model back toward its initialization recovers the robustness that finetuning destroys~\cite{wortsman_robust_2022}. Other work sums task vectors~\cite{ilharco_editing_2023} or weights them by Fisher information~\cite{matena_merging_2022}. 
For image classification, weight averaging combines neural networks trained on successive tasks~\cite{marouf_weighted_2024}. Similarly, in VLA models, interpolation combines a finetuned policy with its pretrained generalist to recover prior capabilities~\cite{yadavRobustFinetuningVisionLanguageAction2025}.
All of these methods merge models after training, with weights that do not adapt during finetuning.
Instead, we constrain model parameters during training, weighted by how much each parameter matters to each task, so the policy ends up balancing knowledge acquisition and retention.

Rather than merging models after training, regularization-based methods constrain the parameters during it, penalizing deviation from the previous parameters and weighting that penalty per parameter by an importance estimate taken from the diagonal Fisher Information Matrix~\cite{kirkpatrick_overcoming_2017}, the sensitivity of the network output~\cite{aljundiMemoryAwareSynapses2018}, or the path integral of the loss along training~\cite{zenke_continual_2017}, while LwF~\cite{li_learning_2016} regularizes the function instead by distilling the old model's outputs. This weight is fixed before training, so it commits to a single trade-off between learning and remembering. We keep the per-parameter importance but turn the penalty into a constraint, solved with the method of multipliers, so its strength grows while the policy is still drifting.

\section{Problem Formulation}\label{sec:problem}
In this section, we formalize the problem of continual learning for finetuning a VLA. We consider a policy finetuned on tasks one at a time. Starting from the pretrained model, we finetune on the first task and keep the resulting parameters as the starting point for the second, and so on. So, each task is learned from the optimal policy of the previous task, and no earlier demonstrations are revisited. This is how a deployed robot accumulates skills. It is also where forgetting compounds, since every finetuning run moves the parameters further from the values the earliest tasks rely on.

Formally, we model each task as a Partially Observable Markov decision process (POMDP) with state space $\mathcal{S}$, action space $\mathcal{A}$, and observation space $\mathcal{O}$. The policy does not have access to the state $s_t \in \mathcal{S}$; at each timestep it instead receives an observation $o_t \in \mathcal{O}$, whose elements include images and proprioception. Since we learn from demonstrations, the transition and reward components of the POMDP play no role and we omit them. We write $\mathcal{T}$ for the space of task descriptions, given as language prompts. A policy $\pi_\theta(a_t|o_t,T)$, parameterized by the vector of network weights $\theta$, maps the observation $o_t \in \mathcal{O}$ and a task description $T \in \mathcal{T}$ to a distribution over actions $a_t \in \mathcal{A}$. We use a subscript $i$ to index finetuning tasks and a subscript $t$ to index timesteps within an episode. We denote $\theta^*_{i}$ as the optimal model parameters at the end of task $T_i$ finetuning. We want to finetune a model, pretrained on $m$ pretraining tasks, on $n$ new task whose demonstrations are available. We write $\mathcal{D}_i = \{\tau^{(j)}\}_{j=1}^{N_i}$ to denote the demonstration data set of task $i \in \{1,\dots,n\}$, where each demonstration $\tau^{(j)} = (o_t, a_t)_{t=1}^{H_j}$ is a trajectory of $H_j$ observation-action pairs and $N_i$ is the number of demonstrations available for task $i$.

To finetune on one of these new tasks, we use the standard behavior cloning (BC) objective~\cite{bain_framework_1995}, also known as supervised finetuning (SFT). For a policy $\pi_\theta$ parameterized by $\theta$ and a demonstration set $
\mathcal{D}_i$, we define the BC loss for task $T_i$ as
\begin{equation}\label{eq:BCloss}
     L^{BC}(\theta;\mathcal{D}_i) = -\frac{1}{\sum_{j=1}^{N_i} H_j} \sum_{j=1}^{N_i} \sum_{t=1}^{H_j} \log \pi_\theta(a_t \mid o_t, T_i).
\end{equation}
We denote by $\pi_\theta^{n}$ the final policy parameters after sequential finetuning from task $T_1$ to $T_n$. Our goal is to learn $\pi_\theta^{n}$ such that it achieves a high success rate on all $m+n$ tasks. 
Note that we do not assume access to the demonstrations used to pretrain the model, nor to those of any task the policy has already been finetuned on.


\section{Preliminaries}\label{sec:prelims}
In this section, we review some results that we will use in our pipeline. 
We review the method of multipliers~\cite{boyd_distributed_2011} that is commonly used to solve constrained optimization problems. 
Then, in Section~\ref{sec:method}, we will show how the continual learning formulation~\eqref{eq:BCloss} from Section~\ref{sec:problem} can be cast as a constrained optimization problem and solved via the method of multipliers.

Consider a constrained optimization problem with cost function $L(\phi)$ and constraint function $g(\phi)$ defined as
\begin{equation}\label{eq:primal_problem}
    \min_{\phi} L(\phi) \hspace{5mm} \text{s.t.} \hspace{5mm} g(\phi) = 0,
\end{equation}
where $\phi$ represents the decision variable. To solve \eqref{eq:primal_problem}, we form the augmented Lagrangian $\mathcal{L}_a$, which adds to the objective a linear term that prices the current constraint violation at the dual variable $u$ and a quadratic term that penalizes the violation regardless of sign,
\begin{equation}\label{eq:aug_lag}
    \mathcal{L}_a := L(\phi) + g(\phi)'u + \frac{\rho}{2}||g(\phi)||^2_2,
\end{equation}
where $g(\phi)'$ is the transpose of $g(\phi)$, $u$ is the vector of dual variables, and $\rho > 0$ is a scalar penalty hyperparameter that sets how strongly the augmented Lagrangian penalizes a violated constraint. The dual variable brings the constraint violation into the objective cost through the term $g(\phi)'u$, so the larger $u$ becomes, the more a violation costs. We can minimize the augmented Lagrangian with the method of multipliers, which iteratively minimizes \eqref{eq:aug_lag} by applying dual ascent~\cite{boyd_distributed_2011}. Let the superscript $k$ denote the iteration index. We solve~\eqref{eq:aug_lag} through the following:
\begin{subequations}\label{eq:method_multiplier}
\begin{align}
\phi^k &= \arg\min_{\phi} L(\phi) + g(\phi)'u^{k-1} + \tfrac{\rho}{2}\|g(\phi)\|^2_2 \label{eq:primal}, \\
u^k      &= u^{k-1} + \rho\, g(\phi^k). \label{eq:dual}
\end{align}
\end{subequations}
The update in \eqref{eq:primal} is the primal update and the update in \eqref{eq:dual} is the dual update. The dual variable $u$ accumulates the violation $g(\phi)$ across iterations and feeds it back into the primal objective, so the longer a violation persists, the higher the augmented objective cost becomes. A fixed penalty has no such memory and settles at a compromise; here the price keeps rising until the constraint is satisfied. The penalty is therefore dynamic rather than static: it adjusts itself as the optimization proceeds instead of being fixed in advance.

\section{\methodname}\label{sec:method}

In this section, we present \methodname, which casts continual learning as a constrained optimization problem. We show how the resulting optimization problem can be solved by the method of multipliers, which constrains the policy's parameters to stay close to the values the earlier tasks rely on. We then make the constraint selective by importance weighting to relax it and allow the parameters to learn the new task.
Unlike existing regularization-based methods that rely solely on a quadratic penalty, \methodname\ incorporates a dual variable term alongside importance weighting. Our empirical results demonstrate that this design effectively balances new task acquisition with prior knowledge retention.

\subsection{Continual Learning as Constrained Optimization Problem}
When learning to do a new task, we assume that policy must minimize the BC loss for the new task subject to the constraint that its parameters do not diverge from the previous task's optimal parameters. We denote $\theta^*_{i-1}$ as the parameters that perform well on the $m$ pretraining tasks together with the $i-1$ finetuning tasks completed so far. We write our optimization problem for task $T_i$ as
\begin{equation} \label{eq:method_primal}
    \min_{\theta_i} L^{BC}(\theta_i;\mathcal{D}_i) \hspace{5mm} \text{s.t.} \hspace{5mm} \theta_i = \theta^*_{i-1},
\end{equation}
where we define $L^{BC}(\theta_i;\mathcal{D}_i)$ in~\eqref{eq:BCloss}. 
To solve~\eqref{eq:method_primal} by the method of multipliers, the primal and dual updates at optimization step $k$ become
\begin{subequations}\label{eq:taco_hard}
\begin{align}
\theta^k_i &= \arg\min_{\theta_i} L^{BC}(\theta_i;\mathcal{D}_i) + (\theta_i - \theta^*_{i-1})'u^{k-1} \nonumber \\
&\quad + \tfrac{\rho}{2}\|\theta_i - \theta^*_{i-1}\|^2, \label{eq:taco_hard_primal} \\
u^k      &= u^{k-1} + \rho\, (\theta_i^k - \theta^*_{i-1}). \label{eq:taco_hard_dual}
\end{align}
\end{subequations}

This penalizes the drift of the model parameters without requiring demonstrations from the previous tasks. As we will show in Section~\ref{sec:sim_results}, this constraint may be too strong to the point that the model does not forget the previous task, but it also cannot learn the new one. For this reason, we selectively choose the parameters to relax the constraint of forgetting and allow the parameters to learn the new task.
\vspace{-5pt}
\subsection{Importance Weighted Anchor}
The constraint in~\eqref{eq:method_primal} treats every parameter alike, making the constraint rigid. To relax the constraint, we hold the parameters that matter most to the earlier tasks near their previous values and leave the rest free to move, so the policy can learn the new task. We do this by replacing the constraint  $\theta_i = \theta^*_{i-1}$ in~\eqref{eq:method_primal} with anchor $z$, an average of $\theta_i$ and  $\theta^*_{i-1}$ weighted by how important each parameter is to each set of tasks. Then, at optimization step $k$, the anchor $z$ is given by
\begin{equation}\label{eq:wtaco_z}
\begin{aligned}
z^k &:= (\Omega_i+ \bar{\Omega}_{i-1})^{-1}(\Omega_i\theta_i^k + \bar{\Omega}_{i-1}\theta^*_{i-1}),
\end{aligned}
\end{equation}
where $\bar{\Omega}_{i-1} \in \mathbb{R}^{d \times d}$, with $d$ the number of policy parameters. The importance matrix $\bar{\Omega}_{i-1}$ is a diagonal matrix that holds one value per parameter measuring its importance to all $m$ pretraining tasks and $i-1$ finetuning tasks that the policy has already learned, and $\Omega_i \in \mathbb{R}^{d \times d}$ measures the importance of each parameter to the new task $T_i$. We denote $\bar{\Omega}_{i}$ as the importance matrix for all $m+i$ tasks and $\Omega_{i}$ as the importance matrix for $T_i$ task. A parameter that matters far more to the prior tasks than to the new one is pulled almost entirely toward $\theta^*_{i-1}$, and the constraint holds it in place. A parameter that matters mainly to the new task is pulled toward its own current value $\theta_i^k$, so the constraint does not pull it toward its older value.

To determine the weight on each parameter in $\theta_i$, note that a parameter whose gradient is large on a task's data is generally the one that matters more to the target task. We therefore estimate the importance $\Omega_i$ of each parameter for task $T_i$ by the mean squared gradient of the task loss over the $N_B$ samples in a batch,
\begin{equation}\label{eq:importance}
 \Omega_i =  \mathrm{diag}\Big( \frac{1}{N_B}\sum_{j=1}^{N_B} \nabla_{\theta_i} L^{BC}(\theta_i; b_j)\,\big(\nabla_{\theta_i} L^{BC}(\theta_i; b_j)\big)'\Big),
\end{equation}
where $b_1,\dots,b_{N_B}$ are the individual samples of a single batch drawn from $\mathcal{D}_i$. This estimate is the diagonal of the empirical Fisher Information Matrix (FIM), the same quantity Elastic Weight Consolidation
(EWC)~\cite{kirkpatrick_overcoming_2017} uses to weight its penalty. We use only diagonal of FIM as the Importance Matrix because it can be easily batched and is more compute and memory efficient as shown in Algorithm~\ref{alg:wtaco}. We can now use the importance estimate of~\eqref{eq:importance} in the anchor~\eqref{eq:wtaco_z} and enforce that anchor as our constraint in~\eqref{eq:taco_hard}, which results in our proposed algorithm \methodname. At optimization step $k$, the primal, dual, and anchor updates are
\begin{equation}\label{eq:wtaco}
\begin{aligned}
\theta^k_i &= \arg\min_{\theta_i} L^{BC}(\theta_i;\mathcal{D}_i)
              + (\theta_i - z^{k-1})'u^{k-1} \\
           &\quad + \tfrac{\rho}{2}\|\theta_i - z^{k-1}\|^2 \\
u^k        &= u^{k-1} + \rho\,(\theta_i^k - z^{k-1}) \\
z^k &= (\Omega_i+ \bar{\Omega}_{i-1})^{-1}(\Omega_i\theta_i^k + \bar{\Omega}_{i-1}\theta^*_{i-1}).
\end{aligned}
\end{equation}

Once finetuning on task $T_i$ finishes, $T_i$ itself becomes a task the policy must not forget. We therefore combine its importance $\Omega_i$ into the old tasks' importance matrix, weighting each term by the number of tasks it covers,
\begin{equation} \label{eq:omega_update}
    \bar{\Omega}_{i} := \dfrac{(m+i-1)\,\bar{\Omega}_{i-1} + \Omega_i}{m+i},
\end{equation}
so that when the policy moves on to $T_{i+1}$, the anchor holds $T_i$ in place along with everything learned before it. Repeating this at every task is what allows \methodname\ to keep learning sequentially.

We provide the pseudocode of our algorithm in Algorithm~\ref{alg:wtaco}. 
Note that we never recompute $\bar{\Omega}_{i-1}$ from the earlier tasks, as we assume previous demonstration data are unavailable. We assume access to only to an importance matrix $\bar{\Omega}_0$ released with the pretrained model. Therefore, we start from $\bar{\Omega}_0$, released with the pretrained model, and combine each pretrained task's importance into the average as that task finishes. This costs one matrix of fixed size, regardless of the number of tasks the policy learns. Such a statistic is computed once during pretraining and is cheaper to store and distribute than the pretraining demonstrations themselves.

\begin{algorithm}[t]
\caption{\methodname}
\label{alg:wtaco}
\begin{algorithmic}[1]
\Require Pretrained VLA $\pi^{\mathrm{pre}}_\theta$ with parameters $\theta^*_0$ ($m$ pretraining tasks); pretraining importance matrix $\bar{\Omega}_0$; demonstration sets $\{\mathcal{D}_i\}$ for task $i \in \{1,..,n\}$; constraint penalty hyperparameter $\rho > 0$; total steps $K$; learning rate $\eta$; number of minibatches $N_B$; stabilizer $\epsilon$
\Ensure $\pi^n_\theta$ has high success rate on all $m+n$ tasks
\For{$i = 1$ \textbf{to} $n$}
  \State $\theta \gets \theta^*_{i-1}$; \quad $u \gets 0$
  \State $\Omega_i \gets \textsc{Importance}(\theta^*_{i-1}, \mathcal{D}_i)$
  \State $z \gets \dfrac{\bar{\Omega}_{i-1} \odot \theta^*_{i-1}
                        + (\Omega_i + \epsilon) \odot \theta}
                       {\bar{\Omega}_{i-1} + \Omega_i + \epsilon}$
  \For{$k = 1$ \textbf{to} $K$}
    \State sample minibatch $b \sim \mathcal{D}_i$
    \State $g \gets \nabla_\theta L^{BC}(\theta; b) + u + \rho\,(\theta - z)$
      \Comment{$\nabla_\theta \mathcal{L}_a$}
    \State $\theta \gets \textsc{OptimStep}(\theta, g, \eta)$
      \Comment{AdamW~\cite{loshchilov_decoupled_2019}} 
      \State $u \gets u + \rho\,(\theta - z)$
      \Comment{dual ascent}
    \State $z \gets \dfrac{\bar{\Omega}_{i-1} \odot \theta^*_{i-1}
                          + (\Omega_i + \epsilon) \odot \theta}
                         {\bar{\Omega}_{i-1} + \Omega_i + \epsilon}$
  \EndFor
  \State $\theta^*_{i} \gets \theta$
  \State $\bar{\Omega}_{i} \gets \dfrac{(m+i-1)\,\bar{\Omega}_{i-1} + \Omega_i}{m+i}$
\EndFor
\State \Return $\pi^n_\theta$ with $\theta^*_{n}$
\Statex
\Function{Importance}{$\theta, \mathcal{D}$}
  \State $\Omega \gets 0$
  \For{$j = 1$ \textbf{to} $N_B$}
    \State sample minibatch $b \sim \mathcal{D}$
    \State $\Omega \gets \Omega + 
           \big(\nabla_\theta L^{BC}(\theta; b)\big)^{\odot 2}$
      \Comment{elementwise square}
  \EndFor
  \State \Return $\Omega / N_B$ 
\EndFunction
\end{algorithmic}

\end{algorithm}
\vspace{-10pt}


\section{Simulation Results}\label{sec:sim_results}


In this section, we present our simulation experiments. The goal is to keep learning new tasks without forgetting any of the prior tasks. Single-task finetuning is the precondition for that: a method that already forgets after one task cannot learn more tasks sequentially. We therefore report the single-task experiments first, and then turn to the sequential setting that motivates the work. 

We use pi0.5~\cite{pi05_2025} as our base pretrained policy $\pi_\theta^{pre}$ as it is trained on heterogeneous sources, including demonstrations from multiple robot platforms, web data, and high-level semantic subtask predictions, which let it generalize well on different manipulation environments. The base model pi0.5 cannot perform the tasks in the simulation environment out of the box, so we first pretrain it on a set of tasks in the environment. This grounds the policy in the environment and gives us a set of tasks against which we can measure how much the policy forgets after finetuning.

\subsection{Experimental setup}
We conduct our experiments on LIBERO~\cite{libero_2023}, a benchmark designed for studying knowledge transfer in multitask and lifelong robot learning problems. We evaluate \methodname\ in three settings. The first two finetune on a single task: the \emph{broadly pretrained} regime starts from a policy pretrained on 117 tasks, and the \emph{narrowly pretrained} regime from one pretrained on 32. By comparing these two pretraining regimes, we isolate the effect of pretraining breadth on how much a policy forgets. In the third setting, we sequentially finetune on three tasks, which is closer to practice, where a deployed policy must keep learning new tasks.

In the \emph{broadly pretrained} regime, we pretrain pi0.5 base model on 117 tasks drawn from LIBERO-90 together with the LIBERO-Spatial, LIBERO-Object, and LIBERO-Goal suites, and then perform single-task finetuning individually across three distinct LIBERO-Long target tasks: ``turn on the stove and put the moka pot on it'' (\texttt{pot-on-stove}), ``put the white mug on the left plate and put the yellow and white mug on the right plate'' (\texttt{mugs-on-plates}) and ``put both the alphabet soup and the cream cheese box in the basket'' (\texttt{items-into-basket}). This regime mirrors the setting in which parameter merging baseline RETAIN has been studied for VLA finetuning~\cite{yadavRobustFinetuningVisionLanguageAction2025}. 

In the \emph{narrowly pretrained} regime, we pretrain pi0.5 base model on 32 of the 40 tasks spanning the LIBERO-Spatial, LIBERO-Object, LIBERO-Goal, and LIBERO-10 suites. We then conduct single-task finetuning on two different tasks: ``turn on the stove and put the moka pot on it'' (\texttt{pot-on-stove}) and ``put the black bowl in the bottom drawer of the cabinet and close it'' (\texttt{bowl-in-drawer}). 

\begin{figure}[t]
    \vspace*{4pt}
    \centering
        \includegraphics[width= 0.42\textwidth]{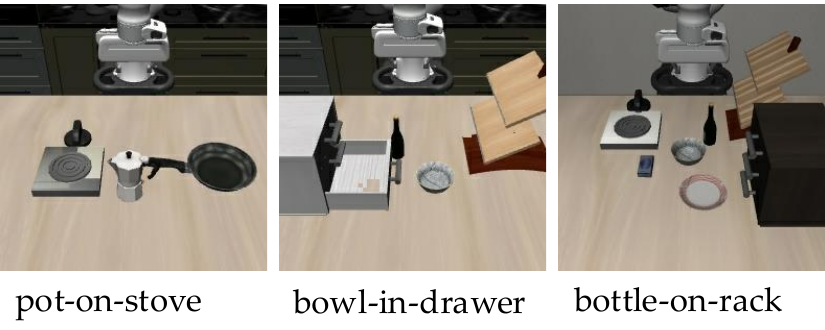}
        \caption{\textbf{The three finetuning LIBERO tasks we learn in sequence.} Starting from a policy pretrained on 32 LIBERO tasks, we finetune on \texttt{pot-on-stove}, then \texttt{bowl-in-drawer}, then \texttt{bottle-on-rack}, in the order shown left to right. Each panel shows the initial scene the policy is given. The first two are long-horizon tasks, while the third is a goal task.}
    \vspace{-20pt}
    \label{fig:libero_task}
\end{figure}

We next examine how \methodname\ behaves under \emph{sequential continual learning}, where we finetune on more than one task in turn. The final policy should perform every task it has been finetuned on as well as the pretraining tasks. We therefore pretrain the pi0.5 base model on 32 tasks and finetune first on \texttt{pot-on-stove} task, then on \texttt{bowl-in-drawer}, and then on ``put the wine bottle on the rack'' (\texttt{bottle-on-rack}) task, as shown in Figure~\ref{fig:libero_task}.
 
\subsection{Evaluation Metric}

We report three success rates: \textbf{Pretrain SR} is the success rate over all the pretraining tasks, and it measures how much of the policy's original knowledge is retained; \textbf{Finetune SR} is the success rate on the finetuning task, and it measures new knowledge acquisition;
\textbf{Weighted Avg SR} weights the pretraining success rate and finetuning success rate by how many tasks each covers, $(N_{\mathrm{pre}}\,SR_{\mathrm{pre}} + N_{\mathrm{ft}}\,{SR}_{\mathrm{ft}})/(N_{\mathrm{pre}}+N_{\mathrm{ft}})$, where $N_{\mathrm{pre}}$ and $N_{\mathrm{ft}}$ are the numbers of pretraining and finetune tasks and $SR$ denotes the corresponding success rates. It is the mean success rate a user would see across every task that the policy is expected to perform, and because pretraining tasks vastly outnumber finetuning tasks, it is typically dominated by retention. We report all the success rate in percent averaged over 5 seeds, 5 episodes/seed for experiments with policy pretrained on 117 tasks (\emph{broadly pretrained}) and 5 seeds, 10 episodes/seed for experiments with policy pretrained on 32 tasks (\emph{narrowly pretrained}) and reported with a $95\%$ confidence interval. 

\subsection{Baselines}
We compare \methodname\ against five replay-free finetuning baselines. \textbf{SFT}~\cite{bain_framework_1995} finetunes on the target task with no mechanism for protecting earlier knowledge, and serves as our reference without any protection against forgetting. \textbf{RETAIN} ~\cite{yadavRobustFinetuningVisionLanguageAction2025} linearly interpolates the finetuned weights with the pretrained weights at a ratio $\alpha$. We use $\alpha = 0.9$ since it was shown to be the most effective hyperparameter on the LIBERO benchmark. \textbf{LoRA}~\cite{lora_2022} freezes the pretrained weights and learns a low-rank update that limits policy drifting by restricting the parameter subspace that the policy can update in. \textbf{EWC}~\cite{kirkpatrick_overcoming_2017} adds a Fisher-weighted L2 regularization selectively pulling parameters toward their pretrained values. \textbf{Simple Recipe Works}~\cite{hu_simple_2026} sequentially adapts the policy to new tasks using on-policy reinforcement learning to update LoRA parameters while keeping the pretrained backbone frozen.
Unlike the supervised baselines, it learns from environment interaction and reward feedback. We also compare against \textbf{\methodname\ w/o importance}, which enforces our anchor with the method of multipliers but the constraint is the previous parameters, as in~\eqref{eq:taco_hard}, and is therefore the ablation of our importance weighting. \textbf{Pretrained (no finetuning)} is the pretrained model without any finetuning and serves as the reference for pretraining-task success before any finetuning. Finally, \textbf{Co-Training}~\cite{dass_datamil_2025} is a replay-based supervised finetuning baseline that samples training examples equally from the current task's demonstrations and a pool of previous-task demonstrations. It acts as a upper-bound reference which finetuning can achieve with access to historical data that \textbf{\methodname} does not require. 

In the broadly pretrained regime, we focus on replay-free supervised finetuning methods. We reserve comparisons with \textbf{Simple Recipe Works} and \textbf{Co-Training} for the narrowly pretrained setting, where more pronounced forgetting provides a stronger test of the trade-off between knowledge acquisition and retention. These additional comparisons assess approaches that use environment interaction and reward feedback, or access to previous demonstrations, respectively.
\subsection{Training details}
In the broadly pretrained regime, we finetune with a batch size of 64 and a learning rate of $2.5 \times 10^{-5}$ for 500 iterations on \texttt{pot-on-stove} and \texttt{items-into-basket} and for 1000 iterations on \texttt{mugs-on-plates}, which reproduces the training setup under which RETAIN was evaluated~\cite{yadavRobustFinetuningVisionLanguageAction2025}. In the narrowly pretrained and sequential finetuning regimes, we finetune every baseline with a batch size of 16 and a learning rate of $1 \times 10^{-5}$ for 10k iterations per task, a smaller batch that the memory of our available GPUs required. We train \methodname\ at the same batch size and learning rate but for 20k iterations per task, because dual ascent raises the cost of violating the anchor only gradually and \methodname\ needs the additional iterations for the constraint to take hold. For the penalty-based baselines we use the same regularisation strength as our method: $\lambda_{\mathrm{EWC}} = \lambda_{\mathrm{L2}} = \rho = 0.02$, with the Fisher diagonal estimated from $N_B = 50$ minibatches at the start of each stage. \textbf{Simple Recipe Works} does not share same iteration budget above because the RL baseline optimises through environment rollouts rather than demonstrations. We adopt the hyperparameters of~\cite{hu_simple_2026}. Since the narrowly pretrained checkpoint attains 0-5\% success on the held-out tasks and therefore yields no reward signal, each stage is preceded by a 100-iteration LoRA behaviour-cloning warm-up on that task's demonstrations. We use four A100 GPUs for finetuning the broadly pretrained policy and 2 A100 GPUs for finetuning the narrowly pretrained policy. Within each regime, we report the best-performing checkpoint of every method.

\subsection{Learning from a broadly pretrained policy}
As shown in Table~\ref{tab:broad}, most of the baselines and \methodname\ reach similar success rates on both the pretraining and the finetuning tasks. Pretraining on this many tasks leaves the policy general enough that it does not forget catastrophically while learning a new one. Even so, \methodname\ has the highest Weighted Avg SR score on two of the three tasks. On \texttt{pot-on-stove}, it retains the most of any method, with the pretraining task success rate dropping by only $1.3\%$ while reaching $92.0\%$ on the finetuning task. On \texttt{items-into-basket}, it retains $94.3\%$ and acquires the task at $100.0\%$ success rate. On \texttt{mugs-on-plates}, LoRA performs better than our method with retention $91.9\%$ against our $90.7\%$, and acquires the task perfectly while we reach $92.0\%$ success rate. Since forgetting is less in this regime, we did not compare against RL-based method and co-training. We turn next to a narrowly pretrained policy, where catastrophic forgetting is more pronounced.


\begin{table}[tb]
\vspace*{4pt}
\centering
\footnotesize
\setlength{\tabcolsep}{4pt}
\begin{tabular}{l ccc}
\toprule
Method & \makecell{Pretrain\\SR $\uparrow$} & \makecell{Finetune\\SR $\uparrow$} & \makecell{Weighted\\Avg SR $\uparrow$}  \\
\midrule
\multicolumn{4}{l}{\emph{Finetune task: \texttt{pot-on-stove}}} \\
\midrule
\textcolor{mygray}{Pretrained (no finetuning)} & \textcolor{mygray}{93.2 $\pm$1.2} & \textcolor{mygray}{4.0 $\pm$11.1} & \textcolor{mygray}{-}  \\ 
SFT & 89.0 $\pm$0.9 & 100.0 $\pm$0.0 & 89.1  \\
RETAIN~\cite{yadavRobustFinetuningVisionLanguageAction2025} & 89.7 $\pm$1.2 & 100.0 $\pm$0.0 & 89.8  \\
LoRA~\cite{lora_2022} & 88.1 $\pm$0.6 & 100.0 $\pm$0.0 & 88.2  \\
EWC~\cite{kirkpatrick_overcoming_2017} & 89.5 $\pm$1.4 & 100.0 $\pm$ 0.0 &  89.6 \\
L2 & 84.8 $\pm$0.7 & 84.0 $\pm$11.1 & 84.8  \\
\methodname\ w/o importance & 89.7 $\pm$1.4 & 56.0 $\pm$32.4 & 89.5  \\
\methodname\ (ours) & 91.9 $\pm$0.9 & 92.0 $\pm$13.6 & \textbf{91.9}  \\
\midrule
\multicolumn{4}{l}{\emph{Finetune task: \texttt{items-into-basket}}} \\
\midrule
\textcolor{mygray}{Pretrained (no finetuning)} & \textcolor{mygray}{93.2 $\pm$1.2} & \textcolor{mygray}{0.0 $\pm$0.0} & \textcolor{mygray}{-}  \\ 
SFT & 93.2 $\pm$1.5 & 100.0 $\pm$0.0 & 93.3  \\
RETAIN~\cite{yadavRobustFinetuningVisionLanguageAction2025} & 93.3 $\pm$2.1 & 96.0 $\pm$11.1 & 93.4  \\
LoRA~\cite{lora_2022} & 93.4 $\pm$0.9 & 100.0 $\pm$0.0 & 93.5  \\
EWC~\cite{kirkpatrick_overcoming_2017} & 93.5 $\pm$1.2 & 100.0 $\pm$0.0 & 93.6  \\
L2 & 93.8 $\pm$1.2 & 92.0 $\pm$13.6 & 93.8  \\
\methodname\ w/o importance & 94.3 $\pm$0.5 & 36.0 $\pm$11.1 & 93.8  \\
\methodname\ (ours) & 94.3 $\pm$1.3 & 100.0 $\pm$0.0 & \textbf{94.4}  \\
\midrule
\multicolumn{4}{l}{\emph{Finetune task: \texttt{mugs-on-plates}}} \\
\midrule
\textcolor{mygray}{Pretrained (no finetuning)} & \textcolor{mygray}{93.2 $\pm$1.2} & \textcolor{mygray}{0.0 $\pm$0.0} & \textcolor{mygray}{-}  \\ 
SFT & 80.7 $\pm$2.0 & 96.0 $\pm$11.1 & 80.8  \\
RETAIN~\cite{yadavRobustFinetuningVisionLanguageAction2025} & 85.4 $\pm$1.1 & 88.0 $\pm$22.2 & 85.5  \\
LoRA~\cite{lora_2022} & 91.9 $\pm$1.3 & 100.0 $\pm$0.0 & 91.9  \\
EWC~\cite{kirkpatrick_overcoming_2017} & 78.8 $\pm$0.8 & 96.0 $\pm$11.1 & 78.9  \\
L2 & 86.1 $\pm$0.6 & 80.0 $\pm$30.4 & 86.0  \\
\methodname\ w/o importance & 93.2 $\pm$1.2 & 76.0 $\pm$20.8 & \textbf{93.1}  \\
\methodname\ (ours) & 90.7 $\pm$1.1 & 92.0 $\pm$13.6 & 90.8  \\
\bottomrule
\end{tabular}
\caption{\textbf{Learning one new task from a broadly pretrained policy.} We finetune a pi0.5 policy pretrained on 117 LIBERO tasks on finetune tasks, and evaluate it on the 117 pretraining tasks (\textbf{Pretrain SR}) and on the finetune task (\textbf{Finetune SR}). All entries are success rates in percent, averaged over 5 random seeds with 5 episodes per seed, reported as mean $\pm$ 95\% confidence interval; higher is better ($\uparrow$) everywhere. \textbf{Weighted Avg SR} is the task-count-weighted mean over all 118 tasks. Forgetting is mild in this regime, and \methodname\ learns each new task without significantly forgetting the 117 pretrained tasks.}
\label{tab:broad}
\vspace{-20pt}
\end{table}

\subsection{Learning from a narrowly pretrained policy}
Table~\ref{tab:narrow} reports the single-task finetuning experiment when the policy has been pretrained on 32 tasks. Forgetting is far more severe here. While most of the baselines learn the finetuning very well and have a high success rate, \methodname\ is the only one able to achieve a high success rate on the pretraining tasks and has a Pretrain SR of approx. $90\%$ whereas all other baselines' Pretrain SR drops below $80\%$ for both tasks. As mentioned in Section~\ref{sec:method}, the \methodname\ without importance is good at retaining knowledge but is inefficient in learning the new task. The importance weighting helps balance knowledge retention and acquisition. 

\begin{table}[tb]
\vspace*{4pt}
\centering
\footnotesize
\setlength{\tabcolsep}{4pt}
\begin{tabular}{l ccc}
\toprule
Method & \makecell{Pretrain\\SR $\uparrow$} & \makecell{Finetune\\SR $\uparrow$} & \makecell{Weighted\\Avg SR $\uparrow$}  \\
\midrule
\multicolumn{4}{l}{\emph{Finetune task: \texttt{pot-on-stove}}} \\
\midrule
\textcolor{mygray}{Pretrained (no finetuning)} & \textcolor{mygray}{93.9 $\pm$1.4} & \textcolor{mygray}{6.0 $\pm$16.7} & \textcolor{mygray}{-}  \\ 
\textcolor{mygray}{Co-Training} &  \textcolor{mygray}{93.8$\pm$1.3} &  \textcolor{mygray}{94.0$\pm$6.8} &  \textcolor{mygray}{\textbf{93.8}} \\
SFT & 56.1 $\pm$2.6 & 96.0 $\pm$6.8 & 57.3  \\
RETAIN~\cite{yadavRobustFinetuningVisionLanguageAction2025} & 66.5 $\pm$2.1 & 94.0 $\pm$11.1 & 67.3  \\
LoRA~\cite{lora_2022} & 75.4 $\pm$1.8 & 100.0 $\pm$0.0 & 76.2  \\
EWC~\cite{kirkpatrick_overcoming_2017} & 47.1 $\pm$1.4 & 96.0 $\pm$11.1 & 48.6  \\
Simple Recipe Works~\cite{hu_simple_2026} & 84.0$\pm$1.3 & 86.0$\pm$16.7 & 84.1  \\
L2 & 71.5 $\pm$2.2 & 74.0 $\pm$16.7 & 71.6  \\
\methodname\ w/o importance & 91.8 $\pm$1.4 & 4.0 $\pm$6.8 & 89.1  \\
\methodname\ (ours) & 89.7 $\pm$1.7 & 86.0 $\pm$18.8 & \textbf{89.6}  \\
\midrule
\multicolumn{4}{l}{\emph{Finetune task: \texttt{bowl-in-drawer}}} \\
\midrule
\textcolor{mygray}{Pretrained (no finetuning)} & \textcolor{mygray}{93.9 $\pm$1.4} & \textcolor{mygray}{0.0 $\pm$0.0} & \textcolor{mygray}{-} \\ 
\textcolor{mygray}{Co-Training} & \textcolor{mygray}{93.1 $\pm$1.7} &  \textcolor{mygray}{100.0$\pm$0.0} & \textcolor{mygray}{\textbf{93.3}}  \\
SFT & 23.1 $\pm$1.4 & 100.0 $\pm$0.0 & 25.4  \\
RETAIN~\cite{yadavRobustFinetuningVisionLanguageAction2025} & 27.4 $\pm$1.8 & 96.0 $\pm$6.8 & 29.5  \\
LoRA~\cite{lora_2022} & 74.0 $\pm$3.0 & 100.0 $\pm$0.0 & 74.8  \\
EWC~\cite{kirkpatrick_overcoming_2017} & 19.6 $\pm$0.9 & 90.0 $\pm$12.4 & 21.7  \\
Simple Recipe Works~\cite{hu_simple_2026} &  74.1$\pm$1.2 &  92.0$\pm$10.4 &  74.6 \\
L2 & 64.9 $\pm$2.3 & 78.0 $\pm$16.2 & 65.3  \\
\methodname\ w/o importance & 90.6 $\pm$1.9 & 38.0 $\pm$30.9 & 89.0  \\
\methodname\ (ours) & 91.8 $\pm$1.2 & 96.0 $\pm$6.8 & \textbf{91.9}  \\
\bottomrule
\end{tabular}
\caption{\textbf{Learning one new task from a narrowly pretrained policy.} We finetune a pi0.5 policy pretrained on 32 LIBERO tasks (8 from each suite) on two LIBERO-Long finetune tasks, and evaluate it on the 32 pretraining tasks (\textbf{Pretrain SR}) and on the finetune task (\textbf{Finetune SR}). All entries are success rates in percent, averaged over 5 random seeds with 10 episodes per seed, reported as mean $\pm$ 95\% confidence interval; higher is better ($\uparrow$) everywhere. \textbf{Weighted Avg SR} is the task-count-weighted mean over all 33 tasks. Forgetting is severe in this regime: every baseline forgets the pretraining tasks, whereas \methodname\ remembers the pretraining tasks.}
\label{tab:narrow}
\vspace{-20pt}
\end{table}

\subsection{Sequential Continual Learning}

As shown in Table~\ref{tab:seq},  \methodname\ is the only method that achieves the right balance between learning the new task and not forgetting the prior task. It still performs \texttt{pot-on-stove} $84.0\%$ of the time and \texttt{bowl-in-drawer} $70.0\%$ of the time after learning a third task on top of them, while retaining $88.2\%$ of the pretraining tasks and learning \texttt{bottle-on-rack} at $82.0\%$. Every baseline succeeds on \texttt{pot-on-stove} exactly $0.0\%$ of the time. All baselines have completely forgotten the first task they learned while still performing the most recent \texttt{bottle-on-rack} task well, with SFT at $84.0\%$ and RETAIN at $96.0\%$ on \texttt{bottle-on-rack}. On the other hand, \methodname\ without importance weighting retains the most of any method at $89.8\%$, but learns almost nothing, ending at $0.0\%$ on the first two tasks and $4.0\%$ on the third. 

\begin{table*}[t]
\vspace*{4pt}
\centering
\footnotesize
\setlength{\tabcolsep}{4pt}
\begin{tabular}{l c ccc c}
\toprule
& & \multicolumn{3}{c}{\textbf{Success on each newly learned task}} & \\
\cmidrule(lr){3-5}
Method & \makecell{Pretrain\\SR $\uparrow$} & \texttt{pot-on-stove} $\uparrow$ & \texttt{bowl-in-drawer} $\uparrow$ & \texttt{bottle-on-rack} $\uparrow$ & \makecell{Weighted\\Avg SR $\uparrow$}  \\
\midrule
\textcolor{mygray}{Co-Training} & \textcolor{mygray}{\textbf{95.4$\pm$1.0}} &  \textcolor{mygray}{\textbf{100.0$\pm$0.0}} &  \textcolor{mygray}{\textbf{98.0$\pm$5.6}} &  \textcolor{mygray}{\textbf{88.0$\pm$13.6}} &  \textcolor{mygray}{\textbf{95.4}} \\
SFT & 12.9 $\pm$1.9 & 0.0 $\pm$0.0 & 60.0 $\pm$21.5 & 84.0 $\pm$11.1 & 15.9  \\
RETAIN~\cite{yadavRobustFinetuningVisionLanguageAction2025} & 17.4 $\pm$0.7 & 0.0 $\pm$0.0 & 68.0 $\pm$16.2 & 96.0 $\pm$6.8 & 20.6  \\
LoRA~\cite{lora_2022} & 45.4 $\pm$1.3 & 0.0 $\pm$0.0 & 18.0 $\pm$13.6 & 92.0 $\pm$10.4 & 44.6  \\
EWC~\cite{kirkpatrick_overcoming_2017} & 0.0 $\pm$0.0 & 0.0 $\pm$0.0 & 0.0 $\pm$0.0 & \textbf{100.0} $\pm$0.0 & 2.9  \\
Simple Recipe Works~\cite{hu_simple_2026} & 44.4$\pm$3.2 & 0.0$\pm$0.0 & 36.0$\pm$14.2 & 90.0$\pm$8.8 &  43.8 \\
L2 & 58.3 $\pm$2.6 & 0.0 $\pm$0.0 & 42.0 $\pm$34.5 & 80.0 $\pm$24.8 & 56.7  \\
\methodname\ w/o importance & \textbf{89.8} $\pm$0.8 & 0.0 $\pm$0.0 & 0.0 $\pm$0.0 & 4.0 $\pm$6.8 & 82.2  \\
\methodname\ (ours) & 88.2 $\pm$2.3 & \textbf{84.0} $\pm$6.8 & \textbf{70.0} $\pm$15.2 & 82.0 $\pm$16.2 & \textbf{87.4}  \\
\bottomrule
\end{tabular}
\caption{\textbf{Sequential Continual Learning.} Starting from the policy pretrained on 32 tasks, we finetune on \texttt{pot-on-stove}, then \texttt{bowl-in-drawer}, then \texttt{bottle-on-rack}. We evaluate the policy after learning all three tasks. All entries are success rates in percent, averaged over 5 random seeds, 10 episodes/seed and reported as mean $\pm$ 95\% confidence interval; higher is better ($\uparrow$) for every column. \textbf{Weighted Avg SR} is the task-count-weighted mean over all 35 tasks. After three tasks every baseline has completely forgotten \texttt{pot-on-stove} task, while \methodname\ still performs it $84.0\%$ of the time along with high success rate on 32 pretrained tasks.}
\label{tab:seq}
\end{table*}

\subsection{Ablation Studies}

SAMBAR combines two mechanisms: a dual variable and an importance-weighted
anchor that decides which parameters the constraint should hold in place. Four of
our methods isolate the two. Every one of them anchors the policy to the parameters
it started the stage from, and all use the same regularisation strength
$\lambda_{\mathrm{EWC}} = \lambda_{\mathrm{L2}} = \rho = 0.02$, so they differ only
in which mechanism is enabled: L2 has neither, EWC adds importance weighting to a
fixed penalty, SAMBAR w/o importance adds dual ascent to a uniform anchor, and
SAMBAR has both.

The effect of dual ascent via method of multipliers while leaving the anchor uniform is seen in results of L2 vs SAMBAR w/o importance. Method of multipliers raises Pretrain SR from $71.5\%$ to $91.8\%$ on \texttt{pot-on-stove}, $64.9\%$ to $90.6\%$ on \texttt{bowl-in-drawer}, and $58.3\%$ to $89.8\%$ in the sequential setting. A fixed penalty settles at a compromise; the dual variable keeps accumulating the violation until the parameters stop drifting. But the dual ascent alone over-constrains the policy and Finetune SR falls from $74.0\%$ to $4.0\%$ on \texttt{pot-on-stove} and from $78.0\%$ to $38.0\%$ on \texttt{bowl-in-drawer}, and in the sequential
setting SAMBAR w/o importance ends at $0.0\%$, $0.0\%$, and $4.0\%$ on the three
finetuning tasks. Holding every parameter equally close to its previous value leaves
no capacity for the new task. 

Adding importance weighting to a fixed penalty which is the case in EWC compared to L2, moves the trade-off the other way: Finetune
SR rises from $74.0\%$ to $96.0\%$ on \texttt{pot-on-stove}, but Pretrain SR collapses
from $71.5\%$ to $47.1\%$, from $64.9\%$ to $19.6\%$ on \texttt{bowl-in-drawer}, and
from $58.3\%$ to $0.0\%$ in the sequential setting. Importance weighting opens room
where the new task needs it, but with a fixed penalty nothing tightens the constraint
back up on the parameters that matter.

The effect of adding importance weighting on top of method of multipliers can be seen in results of SAMBAR w/o importance vs. SAMBAR. SAMBAR recovers acquisition at
almost no cost in retention: Finetune SR goes from $4.0\%$ to $86.0\%$ on
\texttt{pot-on-stove} and from $38.0\%$ to $96.0\%$ on \texttt{bowl-in-drawer}, while
Pretrain SR moves by at most $2.1$ points. This shows how both components are important to balance knowledge acquisition with knowledge retention.

\section{Hardware Results}\label{sec:hardware_results}
\begin{table*}[t]
\centering
\footnotesize
\setlength{\tabcolsep}{4pt}
\begin{tabular}{l cccc ccc}
\toprule
& \multicolumn{4}{c}{\textbf{Pretraining tasks}} & & & \\
\cmidrule(lr){2-5}
Method & \texttt{bird-in-bin} $\uparrow$ & \texttt{two-birds-in-bin} $\uparrow$ & \texttt{eraser-in-mug} $\uparrow$ & \texttt{eraser-in-bowl} $\uparrow$ & \makecell{Pretrain\\SR $\uparrow$} & \makecell{Finetune\\SR $\uparrow$} & \makecell{Weighted\\Avg SR $\uparrow$}  \\
\midrule
\multicolumn{8}{l}{\emph{Finetune task:} \texttt{cake-on-cake-tray}} \\
\midrule
SFT & 4/5 & 1/5 & 5/5 & 2/5 & 60.0 & 5/5 & 68.0  \\
RETAIN ($\alpha=0.5$) & 5/5 & 4/5 & 3/5 & 5/5 & 85.0 & 5/5 & 88.0  \\
EWC & 5/5 & 4/5 & 5/5 & 2/5 & 80.0 & 4/5 & 80.0  \\
\methodname\ (ours) & 5/5 & 5/5 & 5/5 & 5/5 & \textbf{100.0} & 5/5 & \textbf{100.0}  \\
\midrule
\multicolumn{8}{l}{\emph{Finetune task: \texttt{remove-lid-and-place-bird}}} \\
\midrule
SFT & 5/5 & 4/5 & 4/5 & 3/5 & 80.0 & 5/5 & 84.0  \\
RETAIN ($\alpha=0.5$) & 5/5 & 5/5 & 3/5 & 5/5 & 90.0 & 0/5 & 72.0  \\
EWC & 5/5 & 5/5 & 3/5 & 5/5 & 90.0 & 0/5 & 72.0  \\
\methodname\ (ours) & 5/5 & 5/5 & 5/5 & 5/5 & \textbf{100.0} & 4/5 & \textbf{96.0}  \\
\bottomrule
\end{tabular}
\caption{\textbf{Learning one new task on hardware.} We pretrain pi0.5 on four tasks and then finetune it on each of the two finetuning tasks. We evaluate every task with 5 rollouts and report the individual tasks as the number of successful rollouts out of 5. \textbf{Pretrain SR} is the percentage of the 20 pretraining rollouts that succeed, \textbf{Weighted Avg SR} is the percentage of all 25 rollouts that succeed. Higher is better ($\uparrow$) for every column. \methodname\ is the only method that keeps all four pretraining tasks intact while learning the new one.}
\label{tab:hardware}
\end{table*}

\begin{table*}[t]
\centering
\footnotesize
\setlength{\tabcolsep}{4pt}
\begin{tabular}{l cccc cccc}
\toprule
& \multicolumn{4}{c}{\textbf{Pretraining tasks}} & & \multicolumn{2}{c}{\textbf{Finetune tasks}} & \\
\cmidrule(lr){2-5}\cmidrule(lr){7-8}
Method & \makecell{\texttt{bird-in-}\\\texttt{bin} $\uparrow$} & \makecell{\texttt{two-birds-}\\\texttt{in-bin} $\uparrow$} & \makecell{\texttt{eraser-in-}\\\texttt{mug} $\uparrow$} & \makecell{\texttt{eraser-in-}\\\texttt{bowl} $\uparrow$} & \makecell{Pretrain\\SR $\uparrow$} & \makecell{\texttt{cake-on-}\\\texttt{cake-tray} $\uparrow$} & \makecell{\texttt{remove-lid-and-}\\\texttt{place-bird} $\uparrow$} & \makecell{Weighted\\Avg SR $\uparrow$}  \\
\midrule
SFT & 5/5 & 5/5 & 5/5 & 3/5 & 90.0 & 4/5 & 3/5 & 83.3  \\
RETAIN ($\alpha=0.5$) & 5/5 & 5/5 & 4/5 & 4/5 & 90.0 & 5/5 & 0/5 & 76.7  \\
EWC & 5/5 & 5/5 & 5/5 & 5/5 & \textbf{100.0} & 4/5 & 2/5 & 86.7  \\
\methodname\ (ours) & 5/5 & 5/5 & 5/5 & 5/5 & \textbf{100.0} & 4/5 &4/5 & \textbf{93.3}  \\
\bottomrule
\end{tabular}
\caption{\textbf{Sequential Continual Learning on hardware.} Starting from the policy pretrained on the four tasks, we finetune on the \texttt{cake-on-cake-tray} task and then on the \texttt{remove-lid-and-place-bird} task, and evaluate all six tasks after learning the \texttt{remove-lid-and-place-bird} task. Entries follow the same convention as Table~\ref{tab:hardware}, with \textbf{Weighted Avg SR} now the percentage of all 30 rollouts that succeed.}
\label{tab:hardware_seq}
\vspace{-10pt}
\end{table*}

In this section, we ask whether a policy running on real hardware can keep learning new tasks without forgetting the ones it has already learned. We first finetune on a single new task and then on two tasks in sequence.

\subsection{Experiment Setup}
\begin{figure}
    \vspace{-20pt}
    \centerline{
        \includegraphics[width= 0.4\textwidth]{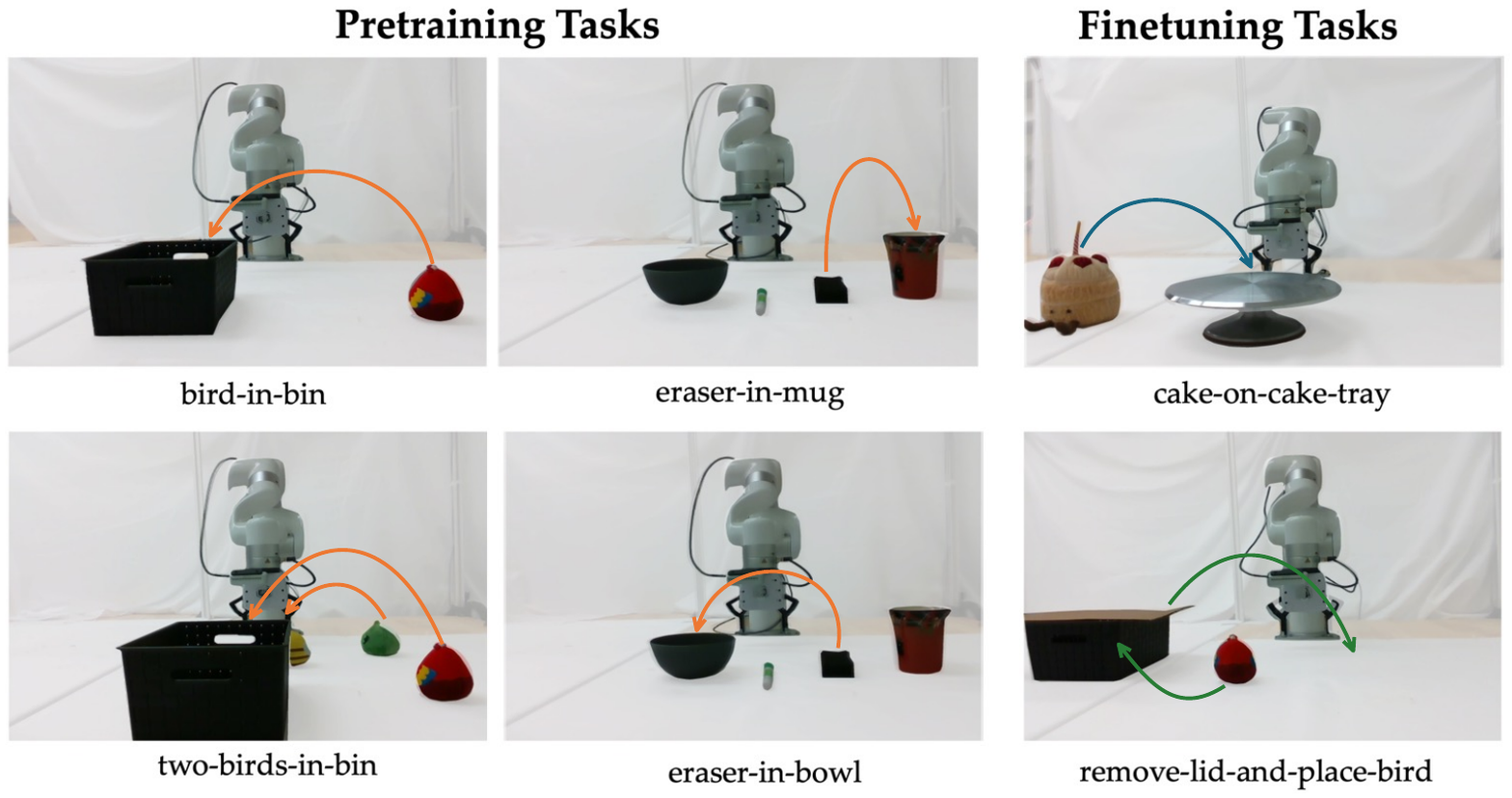}}
        \vspace{-30pt}
        \caption{\textbf{The hardware experiment tasks.} We pretrain pi0.5 on the four tasks on the left and finetune on the two on the right. Each panel shows the initial scene the policy is given and the arrows shows the task policy needs to perform.}
    \vspace{-20pt}
    \label{fig:hardware_task}
\end{figure}

We conduct our hardware experiments on an xArm7 manipulator. We collected 50 tele-operated demonstrations for each of six tabletop tasks shown in Figure~\ref{fig:hardware_task}. 
Four of these serve as pretraining tasks for pretraining the pi0.5 base policy: 
``Pick up red bird and place it in a black bin.'' (\texttt{bird-in-bin}), ``Pick up red bird and green bird and place both in the black bin.'' (\texttt{two-birds-in-bin}), ``Put the green eraser in the red mug.'' (\texttt{eraser-in-mug}) and ``Put the green eraser in a gray bowl.'' (\texttt{eraser-in-bowl}). The remaining two are held out as finetuning tasks: ``put cake on the cake tray'' (\texttt{cake-on-cake-tray}) and ``remove the lid from the bin and place the red bird inside'' (\texttt{remove-lid-and-place-bird}). 
We finetune every method with a batch size of 18 and a learning rate of $1 \times 10^{-5}$ for 10k iterations per task; as in simulation, we give \methodname\ 20k iterations in the sequential setting because of dual ascent it takes longer to achieve similar success rate as SFT.
We report the number of successful rollouts out of 5 for each task, and we count a rollout as successful only if the task is completed in full, awarding no partial credit. We compare against \textbf{SFT}, \textbf{RETAIN} with $\alpha = 0.5$, and \textbf{EWC}, taking one representative of each baseline family from Section~\ref{sec:sim_results}. The evaluation metrics are same as in Section~\ref{sec:sim_results}.

\subsection{Finetuning on a single task}
Table~\ref{tab:hardware} reports results on finetuning on a single task. On \texttt{cake-on-cake-tray}, \methodname\ succeeds on all 5 rollouts of all four pretraining tasks and on all 5 rollouts of the new task, balancing the task of retaining pretraining task knowledge and learning the \texttt{cake-on-cake-tray} task. SFT forgets the long horizon task of \texttt{two-birds-in-bin} and forgets the goal \texttt{eraser-in-bowl} task, where it does not drop the eraser in the bowl. 

The \texttt{remove-lid-and-place-bird} task is a long-horizon task, where RETAIN and EWC both retain the pretraining tasks well, but neither learns the task at all, failing all 5 rollouts. SFT shows the opposite failure: it acquires the finetuning task but forgets the pretraining task. \methodname\ is the only method that keeps the high success rate on all four pretraining tasks while still acquiring the new one.

\subsection{Sequential Continual Learning}

Table~\ref{tab:hardware_seq} reports the results on sequential continual learning on hardware, where we first finetune on \texttt{cake-on-cake-tray} task and then on \texttt{remove-lid-and-place-bird} task. Every method remembers the pretraining tasks, with high pretraining success rate between $90.0\%$ and $100.0\%$, but the methods separate on whether they ever acquire the second task. RETAIN is unable to perform the long horizon \texttt{remove-lid-and-place-bird} task, failing all 5 rollouts, and EWC manages only 2 of 5. \methodname\ succeeds on 4 of 5 rollouts of both finetune tasks while keeping all four pretraining tasks at $100.0\%$, for the best Weighted Avg SR of $93.3\%$ against $86.7\%$ for EWC.


As shown in Sections~\ref{sec:sim_results} and~\ref{sec:hardware_results}, \methodname\ keeps learning new tasks without replaying any demonstration from a task it has already learned. In simulation, every baseline forgets the first task it learned, and on hardware no baseline matches SAMBAR on both retention and learning the second new task. SAMBAR reaches the highest Weighted Avg SR in both settings: 87.4\% in simulation and 93.3\% on hardware.


\section{CONCLUSIONS and FUTURE WORK}\label{sec:conclusion}
We presented SAMBAR, a replay-free method for continual VLA finetuning that combines importance-weighted parameter anchoring with the method of multipliers. On sequential LIBERO tasks, \methodname\ achieves 88.2\% success on pretraining tasks and 78.7\% average success on three newly learned tasks, while every evaluated baseline completely forgets the first finetuning task. On hardware, it retains 100\% success on four pretraining tasks and achieves 80\% on each of two sequentially learned tasks. These results demonstrate the balance \methodname\ provides between knowledge acquisition and retention using stored parameters and importance estimates. Future work will let the policy improve from its own rollouts, updating on the experience it collects without losing the tasks it already performs.




\bibliographystyle{IEEEtran}
\bibliography{references.bib}

@IEEEtranBSTCTL{IEEEexample:BSTcontrol,
    CTLuse_forced_etal       = "yes",
    CTLmax_names_forced_etal = "1",
    CTLnames_show_etal       = "2"
}

@article{dass_datamil_2025,
  title   = {{DataMIL}: Selecting Data for Robot Imitation Learning with Datamodels},
  author  = {Dass, Shivin and Khaddaj, Alaa and Engstrom, Logan and
             Madry, Aleksander and Ilyas, Andrew and
             Mart{\'i}n-Mart{\'i}n, Roberto},
  journal = {arXiv preprint arXiv:2505.09603},
  year    = {2025},

}

@article{kirkpatrick_overcoming_2017,
    title = {Overcoming catastrophic forgetting in neural networks},
    volume = {114},
    issn = {0027-8424, 1091-6490},
    doi = {10.1073/pnas.1611835114},
    number = {13},
    urldate = {2026-02-13},
    journal = {Proceedings of the National Academy of Sciences},
    author = {Kirkpatrick, James and Pascanu, Razvan and Rabinowitz, Neil and Veness, Joel and Desjardins, Guillaume and Rusu, Andrei A. and Milan, Kieran and Quan, John and Ramalho, Tiago and Grabska-Barwinska, Agnieszka and Hassabis, Demis and Clopath, Claudia and Kumaran, Dharshan and Hadsell, Raia},
    month = mar,
    year = {2017},
    pages = {3521--3526},
}

@article{langeContinualLearningSurvey2021,
    title = {A continual learning survey: {Defying} forgetting in classification tasks},
    shorttitle = {A continual learning survey},
    issn = {0162-8828, 2160-9292, 1939-3539},
    doi = {10.1109/TPAMI.2021.3057446},
    journal = {IEEE Transactions on Pattern Analysis and Machine Intelligence},
    shortjournal = {IEEE Trans. Pattern Anal. Mach. Intell.},
    author = {Lange, Matthias De and Aljundi, Rahaf and Masana, Marc and Parisot, Sarah and Jia, Xu and Leonardis, Ales and Slabaugh, Gregory and Tuytelaars, Tinne},
    year = {2021},
    pages = {1--1},
}

@incollection{bain_framework_1995,
    title = {A framework for behavioural cloning},
    booktitle = {Machine {Intelligence} 15},
    author = {Bain, Michael and Sammut, Claude},
    year = {1995},
    pages = {103--129},
}

@inproceedings{libero_2023,
    title = {{LIBERO}: {Benchmarking} {Knowledge} {Transfer} for {Lifelong} {Robot} {Learning}},
    shorttitle = {{LIBERO}},
    doi = {10.48550/arXiv.2306.03310},
    booktitle = {Advances in {Neural} {Information} {Processing} {Systems} ({NeurIPS}) {Datasets} and {Benchmarks} {Track}},
    author = {Liu, Bo and Zhu, Yifeng and Gao, Chongkai and Feng, Yihao and Liu, Qiang and Zhu, Yuke and Stone, Peter},
    year = {2023},
}

@inproceedings{lora_2022,
    title = {{LoRA}: {Low}-{Rank} {Adaptation} of {Large} {Language} {Models}},
    shorttitle = {{LoRA}},
    doi = {10.48550/arXiv.2106.09685},
    booktitle = {International {Conference} on {Learning} {Representations} ({ICLR})},
    author = {Hu, Edward J. and Shen, Yelong and Wallis, Phillip and Allen-Zhu, Zeyuan and Li, Yuanzhi and Wang, Shean and Wang, Lu and Chen, Weizhu},
    year = {2022},
}

@misc{yadavRobustFinetuningVisionLanguageAction2025,
    title = {Robust {Finetuning} of {Vision}-{Language}-{Action} {Robot} {Policies} via {Parameter} {Merging}},
    doi = {10.48550/arXiv.2512.08333},
    urldate = {2026-02-04},
    publisher = {arXiv},
    author = {Yadav, Yajat and Zhou, Zhiyuan and Wagenmaker, Andrew and Pertsch, Karl and Levine, Sergey},
    month = dec,
    year = {2025},
    note = {arXiv:2512.08333},
}

@misc{pi05_2025,
    title = {{$\pi_{0.5}$}: a {Vision}-{Language}-{Action} {Model} with {Open}-{World} {Generalization}},
    shorttitle = {{$\pi_{0.5}$}},
    doi = {10.48550/arXiv.2504.16054},
    urldate = {2026-08-11},
    publisher = {arXiv},
    author = {{Physical Intelligence} and Black, Kevin and Brown, Noah and Darpinian, James and Dhabalia, Karan and Driess, Danny and others},
    month = apr,
    year = {2025},
    note = {arXiv:2504.16054},
}

@misc{liuPretrainedVisionLanguageActionModels2026,
    title = {Pretrained {Vision}-{Language}-{Action} {Models} are {Surprisingly} {Resistant} to {Forgetting} in {Continual} {Learning}},
    doi = {10.48550/arXiv.2603.03818},
    urldate = {2026-07-08},
    publisher = {arXiv},
    author = {Liu, Huihan and Kim, Changyeon and Liu, Bo and Liu, Minghuan and Zhu, Yuke},
    month = mar,
    year = {2026},
    note = {arXiv:2603.03818},
}

@inproceedings{zhouTACOTemporalConsensus2026,
    title = {{TACO}: {Temporal} {Consensus} {Optimization} for {Continual} {Neural} {Mapping}},
    shorttitle = {{TACO}},
    booktitle = {Robotics: {Science} and {Systems} ({RSS})},
    author = {Zhou, Xunlan and Zhao, Hongrui and Mehr, Negar},
    year = {2026},
}

@misc{aljundiMemoryAwareSynapses2018,
    title = {Memory {Aware} {Synapses}: {Learning} what (not) to forget},
    shorttitle = {Memory {Aware} {Synapses}},
    doi = {10.48550/arXiv.1711.09601},
    urldate = {2026-02-17},
    publisher = {arXiv},
    author = {Aljundi, Rahaf and Babiloni, Francesca and Elhoseiny, Mohamed and Rohrbach, Marcus and Tuytelaars, Tinne},
    month = oct,
    year = {2018},
    note = {arXiv:1711.09601},
}

@article{boyd_distributed_2011,
    title = {Distributed optimization and statistical learning via the alternating direction method of multipliers},
    volume = {3},
    issn = {1935-8237},
    doi = {10.1561/2200000016},
    number = {1},
    journal = {Foundations and Trends in Machine Learning},
    author = {Boyd, Stephen and Parikh, Neal and Chu, Eric and Peleato, Borja and Eckstein, Jonathan},
    year = {2011},
    pages = {1--122},
}

@inproceedings{rt2_2023,
    title = {{RT}-2: {Vision}-{Language}-{Action} {Models} {Transfer} {Web} {Knowledge} to {Robotic} {Control}},
    shorttitle = {{RT}-2},
    doi = {10.48550/arXiv.2307.15818},
    booktitle = {Conference on {Robot} {Learning} ({CoRL})},
    author = {Brohan, Anthony and Brown, Noah and Carbajal, Justice and Chebotar, Yevgen and Chen, Xi and Choromanski, Krzysztof and others},
    year = {2023},
}

@inproceedings{openvla_2024,
    title = {{OpenVLA}: {An} {Open}-{Source} {Vision}-{Language}-{Action} {Model}},
    shorttitle = {{OpenVLA}},
    doi = {10.48550/arXiv.2406.09246},
    booktitle = {Conference on {Robot} {Learning} ({CoRL})},
    author = {Kim, Moo Jin and Pertsch, Karl and Karamcheti, Siddharth and Xiao, Ted and Balakrishna, Ashwin and Nair, Suraj and others},
    year = {2024},
}

@inproceedings{lopezpaz_gradient_2017,
    title = {Gradient {Episodic} {Memory} for {Continual} {Learning}},
    doi = {10.48550/arXiv.1706.08840},
    booktitle = {Advances in {Neural} {Information} {Processing} {Systems} ({NeurIPS})},
    author = {Lopez-Paz, David and Ranzato, Marc'Aurelio},
    year = {2017},
}

@misc{chaudhry_tiny_2019,
    title = {On {Tiny} {Episodic} {Memories} in {Continual} {Learning}},
    doi = {10.48550/arXiv.1902.10486},
    publisher = {arXiv},
    author = {Chaudhry, Arslan and Rohrbach, Marcus and Elhoseiny, Mohamed and Ajanthan, Thalaiyasingam and Dokania, Puneet K. and Torr, Philip H. S. and Ranzato, Marc'Aurelio},
    year = {2019},
    note = {arXiv:1902.10486},
}

@inproceedings{wan_lotus_2024,
    title = {{LOTUS}: {Continual} {Imitation} {Learning} for {Robot} {Manipulation} {Through} {Unsupervised} {Skill} {Discovery}},
    shorttitle = {{LOTUS}},
    doi = {10.48550/arXiv.2311.02058},
    booktitle = {{IEEE} {International} {Conference} on {Robotics} and {Automation} ({ICRA})},
    author = {Wan, Weikang and Zhu, Yifeng and Shah, Rutav and Zhu, Yuke},
    year = {2024},
}

@misc{bethune_scaling_2025,
    title = {Scaling {Laws} for {Forgetting} during {Finetuning} with {Pretraining} {Data} {Injection}},
    doi = {10.48550/arXiv.2502.06042},
    publisher = {arXiv},
    author = {Bethune, Louis and Grangier, David and Busbridge, Dan and Gualdoni, Eleonora and Cuturi, Marco and Ablin, Pierre},
    month = feb,
    year = {2025},
    note = {arXiv:2502.06042},
}

@inproceedings{zenke_continual_2017,
    title = {Continual {Learning} {Through} {Synaptic} {Intelligence}},
    doi = {10.48550/arXiv.1703.04200},
    booktitle = {International {Conference} on {Machine} {Learning} ({ICML})},
    author = {Zenke, Friedemann and Poole, Ben and Ganguli, Surya},
    year = {2017},
}

@inproceedings{li_learning_2016,
    title = {Learning without {Forgetting}},
    doi = {10.48550/arXiv.1606.09282},
    booktitle = {European {Conference} on {Computer} {Vision} ({ECCV})},
    author = {Li, Zhizhong and Hoiem, Derek},
    year = {2016},
}

@inproceedings{mallya_packnet_2018,
    title = {{PackNet}: {Adding} {Multiple} {Tasks} to a {Single} {Network} by {Iterative} {Pruning}},
    shorttitle = {{PackNet}},
    doi = {10.48550/arXiv.1711.05769},
    booktitle = {{IEEE}/{CVF} {Conference} on {Computer} {Vision} and {Pattern} {Recognition} ({CVPR})},
    author = {Mallya, Arun and Lazebnik, Svetlana},
    year = {2018},
}

@inproceedings{farajtabar_orthogonal_2020,
    title = {Orthogonal {Gradient} {Descent} for {Continual} {Learning}},
    doi = {10.48550/arXiv.1910.07104},
    booktitle = {International {Conference} on {Artificial} {Intelligence} and {Statistics} ({AISTATS})},
    author = {Farajtabar, Mehrdad and Azizan, Navid and Mott, Alex and Li, Ang},
    year = {2020},
}

@inproceedings{liu_skill_2025,
    title = {Skill {Expansion} and {Composition} in {Parameter} {Space}},
    doi = {10.48550/arXiv.2502.05932},
    booktitle = {International {Conference} on {Learning} {Representations} ({ICLR})},
    author = {Liu, Tenglong and Li, Jianxiong and Zheng, Yinan and Niu, Haoyi and Lan, Yixing and Xu, Xin and Zhan, Xianyuan},
    year = {2025},
}

@misc{zhu_can_2026,
    title = {Can {Vision}-{Language}-{Action} {Models} {Learn} from {Real}-{World} {Data} {Continually} without {Forgetting}?},
    doi = {10.48550/arXiv.2605.26820},
    publisher = {arXiv},
    author = {Zhu, Jiarun and Hong, Yijun and Sun, Xiaoquan and Xu, Zetian and He, Qijun and Chen, Haijier and others},
    month = may,
    year = {2026},
    note = {arXiv:2605.26820},
}

@misc{liu_lifelongrft_2026,
    title = {Towards {Long}-{Lived} {Robots}: {Continual} {Learning} {VLA} {Models} via {Reinforcement} {Fine}-{Tuning}},
    shorttitle = {Towards {Long}-{Lived} {Robots}},
    doi = {10.48550/arXiv.2602.10503},
    publisher = {arXiv},
    author = {Liu, Yuan and Li, Haoran and Tian, Shuai and Qin, Yuxing and Chen, Yuhui and Zheng, Yupeng and Huang, Yongzhen and Zhao, Dongbin},
    month = feb,
    year = {2026},
    note = {arXiv:2602.10503},
}

@misc{zeng_crlvla_2026,
    title = {{CRL}-{VLA}: {Continual} {Vision}-{Language}-{Action} {Learning}},
    shorttitle = {{CRL}-{VLA}},
    doi = {10.48550/arXiv.2602.03445},
    publisher = {arXiv},
    author = {Zeng, Qixin and Zhang, Shuo and Zhang, Hongyin and Wang, Renjie and Zhao, Han and Zhao, Libang and others},
    month = feb,
    year = {2026},
    note = {arXiv:2602.03445},
}

@misc{hu_simple_2026,
    title = {Simple {Recipe} {Works}: {Vision}-{Language}-{Action} {Models} are {Natural} {Continual} {Learners} with {Reinforcement} {Learning}},
    shorttitle = {Simple {Recipe} {Works}},
    doi = {10.48550/arXiv.2603.11653},
    publisher = {arXiv},
    author = {Hu, Jiaheng and Shim, Jay and Tang, Chen and Sung, Yoonchang and Liu, Bo and Stone, Peter and others},
    month = mar,
    year = {2026},
    note = {arXiv:2603.11653},
}

@misc{guo_priorvla_2026,
    title = {{PriorVLA}: {Prior}-{Preserving} {Adaptation} for {Vision}-{Language}-{Action} {Models}},
    shorttitle = {{PriorVLA}},
    doi = {10.48550/arXiv.2605.10925},
    publisher = {arXiv},
    author = {Guo, Xinyu and Xie, Bin and Chai, Wei and Deng, Xianchi and Wang, Tiancai and Wu, Zhengxing and Chen, Xingyu},
    month = may,
    year = {2026},
    note = {arXiv:2605.10925},
}

@misc{huang_breaking_2026,
    title = {Breaking {Lock}-{In}: {Preserving} {Steerability} under {Low}-{Data} {VLA} {Post}-{Training}},
    shorttitle = {Breaking {Lock}-{In}},
    doi = {10.48550/arXiv.2604.23121},
    publisher = {arXiv},
    author = {Huang, Suning and Shao, Jiaqi and Wang, Ke and Chen, Qianzhong and Sun, Jiankai and Guo, Yanjiang and Schwager, Mac and Bohg, Jeannette},
    month = apr,
    year = {2026},
    note = {arXiv:2604.23121},
}

@inproceedings{frankle_linear_2020,
    title = {Linear {Mode} {Connectivity} and the {Lottery} {Ticket} {Hypothesis}},
    doi = {10.48550/arXiv.1912.05671},
    booktitle = {International {Conference} on {Machine} {Learning} ({ICML})},
    author = {Frankle, Jonathan and Dziugaite, Gintare Karolina and Roy, Daniel M. and Carbin, Michael},
    year = {2020},
}

@inproceedings{wortsman_robust_2022,
    title = {Robust fine-tuning of zero-shot models},
    doi = {10.48550/arXiv.2109.01903},
    booktitle = {{IEEE}/{CVF} {Conference} on {Computer} {Vision} and {Pattern} {Recognition} ({CVPR})},
    author = {Wortsman, Mitchell and Ilharco, Gabriel and Kim, Jong Wook and Li, Mike and Kornblith, Simon and Roelofs, Rebecca and others},
    year = {2022},
}

@inproceedings{ilharco_editing_2023,
    title = {Editing {Models} with {Task} {Arithmetic}},
    doi = {10.48550/arXiv.2212.04089},
    booktitle = {International {Conference} on {Learning} {Representations} ({ICLR})},
    author = {Ilharco, Gabriel and Ribeiro, Marco Tulio and Wortsman, Mitchell and Gururangan, Suchin and Schmidt, Ludwig and Hajishirzi, Hannaneh and Farhadi, Ali},
    year = {2023},
}

@inproceedings{matena_merging_2022,
    title = {Merging {Models} with {Fisher}-{Weighted} {Averaging}},
    doi = {10.48550/arXiv.2111.09832},
    booktitle = {Advances in {Neural} {Information} {Processing} {Systems} ({NeurIPS})},
    author = {Matena, Michael and Raffel, Colin},
    year = {2022},
}

@inproceedings{marouf_weighted_2024,
    title = {Weighted {Ensemble} {Models} {Are} {Strong} {Continual} {Learners}},
    doi = {10.48550/arXiv.2312.08977},
    booktitle = {European {Conference} on {Computer} {Vision} ({ECCV})},
    author = {Marouf, Imad Eddine and Roy, Subhankar and Tartaglione, Enzo and Lathuili{\`e}re, St{\'e}phane},
    year = {2024},
}

@inproceedings{loshchilov_decoupled_2019,
    title = {Decoupled {Weight} {Decay} {Regularization}},
    doi = {10.48550/arXiv.1711.05101},
    booktitle = {International {Conference} on {Learning} {Representations} ({ICLR})},
    author = {Loshchilov, Ilya and Hutter, Frank},
    year = {2019},
}

@misc{molmoact_2025,
    title = {{MolmoAct}: {Action} {Reasoning} {Models} that can {Reason} in {Space}},
    shorttitle = {{MolmoAct}},
    doi = {10.48550/arXiv.2508.07917},
    author = {Lee, Jason and Duan, Jiafei and Fang, Haoquan and Deng, Yuquan and Liu, Shuo and Li, Boyang and Fang, Bohan and Zhang, Jieyu and Wang, Yi Ru and Lee, Sangho and Han, Winson and Pumacay, Wilbert and Wu, Angelica and Hendrix, Rose and Farley, Karen and VanderBilt, Eli and Farhadi, Ali and Fox, Dieter and Krishna, Ranjay},
    year = {2025},
    note = {arXiv:2508.07917},
}

\end{document}